# A.I. Robustness: a Human-Centered Perspective on Technological Challenges and Opportunities

ANDREA TOCCHETTI*, LORENZO CORTI*, AGATHE BALAYN*, MIREIA YURRITA, PHILIP LIPPMANN, MARCO BRAMBILLA, and JIE YANG†

Despite the impressive performance of Artificial Intelligence (AI) systems, their robustness remains elusive and constitutes a key issue that impedes large-scale adoption. Robustness has been studied in many domains of AI, yet with different interpretations across domains and contexts. In this work, we systematically survey the recent progress to provide a reconciled terminology of concepts around AI robustness. We introduce three taxonomies to organize and describe the literature both from a fundamental and applied point of view: 1) *robustness by* methods and approaches in different phases of the machine learning pipeline; 2) *robustness for* specific model architectures, tasks, and systems; and in addition, 3) *robustness assessment* methodologies and insights, particularly the trade-offs with other trustworthiness properties. Finally, we identify and discuss research gaps and opportunities and give an outlook on the field. We highlight the central role of humans in evaluating and enhancing AI robustness considering the necessary knowledge humans can provide, and discuss the need for better understanding practices and developing supportive tools in the future.

## 1 INTRODUCTION

Artificial Intelligence (AI) systems have been widely adopted in diverse areas such as medicine [26] or education [111]. Although AI systems show potential and are expected to revolutionize existing workflows by combining human- and non-human skills [20], there is still little insight into how we should deal with the trade-offs of combining human and artificial agency, or the way in which these systems should be assessed and held accountable [70]. Furthermore, concerns about bias [32], inscrutability [12], and vulnerability [98] have also been raised. Consequently, several social actors, like the European High-Level Expert Group, have highlighted the need for socio-political deliberation around the design and governance of AI systems, and have defined principles for Trustworthy AI, i.e., the *Ethics Guidelines for Trustworthy AI* [187].

One of the core principles of Trustworthy AI is robustness [70], defined in Machine Learning (ML) as *the insensitivity of a model's performance to miscalculations of its parameters* [155, 268]. Examples like Tesla's Full Self-Driving mechanism erroneously identifying the moon as a yellow

---

*The authors contributed equally to this research.

†Andrea Tocchetti and Marco Brambilla are with Politecnico di Milano, Email: {andrea.tocchetti, marco.brambilla}@polimi.it; Lorenzo Corti, Agathe Balayn, Mireia Yurrita, Philip Lippmann, and Jie Yang (corresponding author) are with Delft University of Technology, Email: {l.corti, a.m.a.balayn, m.yurritasemperena, p.lippmann, j.yang-3}@tudelft.nl.

---

traffic light,[1] or Autopilot being fooled by stickers placed on the ground,[2] show that AI systems might be susceptible to errors and vulnerable to external attacks. This may result in undesired behavior and decreased performance [250]. Given the application of AI systems in safety-critical areas (e.g., medical diagnosis [22]), it is paramount to design reliable systems, so that they can be properly and safely integrated in the context of use. In response to this need, a growing body of literature focuses on developing and testing robust AI systems. Methodologies towards robust AI have addressed every phase of the ML pipeline, going from data collection and feature extraction, to model training and prediction [250]. Such methodologies have also been applied to a wide range of tasks and application areas, including (but not limited to) image classification [213] and object detection [45] in Computer Vision, or text classification in Natural Language Processing [113].

Considering the increasing efforts devoted to this field within Trustworthy AI, in this paper we seek to analyze the progress made so far and give a *structured* overview of the suggested solutions. Furthermore, we also aim at identifying the areas that have received least attention, highlighting research gaps, and projecting into future research directions. Our work differs from similar efforts in three main ways. (1) As opposed to some previous work [37, 80, 250], we do not limit the scope of our analysis to adversarial attacks. We argue that, as suggested by Drenkow et al. [64] or Shen et al. [196], natural (i.e., non-adversarial) perturbations constitute a common real-world menace that needs further attention. (2) As far as the application area is concerned, and contrary to surveys solely focusing on tasks like Computer Vision [64] or architectures like Graph Neural Networks [196], we do not limit our survey to any technology in particular. We rather conduct our search in a task-agnostic way. Such an approach helps us identify the most prominent trends within the field and compare the differences in effort and interest across applications as part of our survey. (3) Most importantly, we adopt a human-centered perspective for highlighting the technological challenges and opportunities in the field of robust AI. We argue that previous work, which is predominantly algorithm centric, fails to identify the potential of human input when crafting robust algorithmic systems. We also emphasize the need to understand current human-led practices in order to integrate robustness into existing workflows and tools. To this end, we advocate for a multidisciplinary approach and bring insights from human-centered fields, such as explainable AI, crowd computing, or human-in-the-loop machine learning.

We, therefore, make the following contributions:

(1) We give an overview of the main concepts around robust AI. We consolidate the terminology used in this context, disentangling the meaning and scope of different constructs. We pay special attention to identifying the commonalities and differentiating aspects of the used terms.

(2) We systematically summarize 380 papers on robust AI and related concepts and arrange them in three different taxonomies. First, we group papers that improve *robustness by* working on different aspects of the ML pipeline. We identified three main aspects that the selected studies work on: input data, in-model attributes, and model post-processing aspects. Second, we focus on distinct architectures and application areas of robust AI systems and define *robustness for* specific architectures (e.g., Graph Neural Networks), specific tasks (i.e., Natural Language Processing and Cybersecurity), and systems conceived within other fields of Trustworthy AI (i.e., explainable and fairness-aware systems). We focus on these particular architectures, systems, and fields as they have comparatively received little attention in previous surveys despite the importance of robustness as a desired property. Third, we create a taxonomy related to the *assessment* of robust AI systems.

---

[1] https://www.autoweek.com/news/green-cars/a37114603/tesla-fsd-mistakes-moon-for-traffic-light/ (access 13.10.2022)
[2] https://keenlab.tencent.com/en/whitepapers/Experimental_Security_Research_of_Tesla_Autopilot.pdf (access 13.10.2022)

(3) We identify and discuss disparate research efforts in each of the established fields and identify research gaps. Specifically, we make a special in-depth analysis of the opportunities brought by one of the identified research gaps: the absence of human-centered work in existing methodologies. We highlight the multidisciplinary nature of the robust AI field and provide an outlook for future research directions, bringing insights from human-centered fields.

The remainder of the paper is organized as follows. In section 2, we detail the methodology we used for conducting our systematic review. In section 3, we give an overview of the terms that are currently being used in the field of robust AI, which informed our clustering of the literature. We also clarify the definitions that we will use throughout the paper. Then, in section 4, section 5, and section 6, we conduct our survey and generate the aforementioned taxonomy. In section 7, we identify the most prominent research areas, pinpoint fields that require more research efforts, and highlight future research directions. Finally, in section 8, we further develop our discourse on the lack of human-centered approaches to robust AI, before concluding in section 9.

## 2  SURVEY METHODOLOGY: PAPER COLLECTION

In this chapter, we detail the process applied to collecting the final list of articles considered in this literature review. This includes keyword collection and curation, querying multiple databases, de-duplication, manual filtering, tagging, and analysis.

### 2.1  Collecting Papers

*Defining Keywords.* First, we curated the list of keywords to be used for querying articles. We inspected key definitions of robustness and robust AI [39, 84, 172] in the context of Computer Science and organized a preliminary list. We further enriched this list such that it covers aspects related to the trustworthiness of AI systems and to human-centeredness (including human knowledge) given the lack of a common viewpoint on robustness. Table 1 shows the complete list of keywords used.

| Group Name | Keywords |
|---|---|
| Fundamental | Robustness, Robust |
| Scope | Artificial Intelligence, Machine Learning, Neural Network |
| Context | Trustworthy, Stability, Resilience, Reliability, Accountability, Transparency, Reproducibility Accuracy, Confidence, Performance, Design, Adversarial, Unknowns, Noise Human Computation, Human Knowledge, Human-In-the-Loop, Human Interpretation, Knowledge Base, Knowledge, Knowledge Elicitation, Reasoning Explainability, Explanation, Interpretability, Interpretable |

Table 1.  The groups of keywords considered in the data collection process and the corresponding keywords.

*Querying Publication Databases.* Secondly, we queried multiple databases by generating all possible triples of keywords based on the groups we defined. This led to 156 unique search queries structured as a conjunction between the chosen keywords, e.g., "Robustness" AND "Artificial Intelligence" AND "Explainability". Articles have been collected in July 2022 through *Publish or Perish*[3] by querying the following supported bibliographical databases: Google Scholar[4], Scopus [5],

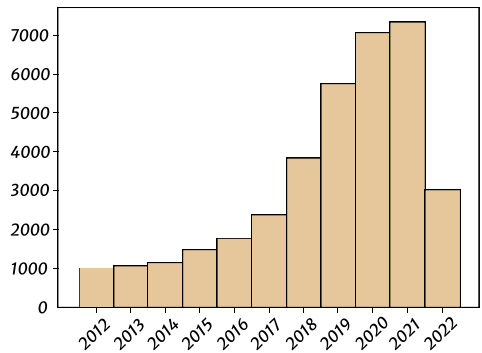

Fig. 1. Temporal distribution of the 35,800 unique papers published in the last 10 years. It is possible to observe a growing trend of published papers about Robust AI over the years. The amount of papers collected in 2022 is not to be considered relevant to this trend as the data was collected in July 2022.

Semantic Scholar [6], and Web of Science [7]. We limited the number of papers collected per query to 200 to comply with the limitations imposed by *Publish or Perish* and the aforementioned bibliographical databases. Moreover, given the breadth of the literature on trustworthy and robust AI, we inspect literature from the last 10 years, i.e., articles published between January 2012 and July 2022.

## 2.2 Filtering Papers

*Pre-filtering.* We collected about 100,000 papers distributed as follows: 31,000 from Google Scholar, 18,450 from Scopus, 30,800 from Semantic Scholar, and 19,400 from Web of Science. Considering the breadth of the data collection, we sought to remove any duplicate entries in our results. Papers that had the same title and authors were filtered out, resulting in 45,400 papers. Duplicates that were undetected at this stage were discarded in the later ones. Then, papers published before the period of interest (January 2012 to July 2022) were filtered out, leading to 35,800 articles. Figure 1 displays the time distribution of the collected papers. We observe a growing interest in the considered topics over the years, which (partially) motivates the time constraints applied.

*Further Inspecting Papers.* At this stage, we manually inspected the abstracts of the collected papers to exclude the ones whose context or content require domain-specific expertise (e.g., healthcare), or deal with a notion of "robustness" that is not related to machine learning (e.g., signal processing). We ended up with 1,800 interesting papers. While inspecting papers, we marked them with specific keywords, e.g., "Computer Vision" or "Loss Function", to differentiate them in terms of content and type of publication (e.g., "Literature Review"). Consequently, we used those keywords to perform a final filtering step in which the papers tagged with the least frequent keywords, i.e., appearing only once, were excluded. Omitted keywords include: "audio signal" and "event detection". Throughout the entire process, we carefully analyzed the papers such that they contain significant or late progress in the area despite them not being peer-reviewed yet (e.g., from arXiv), building our final set of papers. In the end, this thorough inspection led to 560 papers that were systematically analyzed, out of which 380 papers were systematically summarized and discussed[8]. The list of collected, filtered, and summarized papers can be found on GitHub[9].

---

[6]Sematic Scholar: https://www.semanticscholar.org/
[7]Web of Science: https://www.webofscience.com/
[8]Due to space limit, we leave the discussions of some papers (about 30%) in the supplementary material.
[9]https://github.com/AndreaTocchetti/ACMReviewPaperPolimiDelft.git

## 3 OVERVIEW OF THE MAIN CONCEPTS SURROUNDING ROBUSTNESS

From our collection of papers, we evinced that the notion of *Robustness* is ill-defined. A number of Machine Learning sub-domains refer to robustness from different viewpoints. We clarify the relations between these domains in subsection 3.1. We also identify that a number of concepts directly related to robustness are used in different ways across research papers (Figure 2). We disambiguate the interpretation of related terms in subsection 3.2. Finally, our analysis of the papers surfaced a few recurring themes, introduced in subsection 3.3, and used to organize our survey.

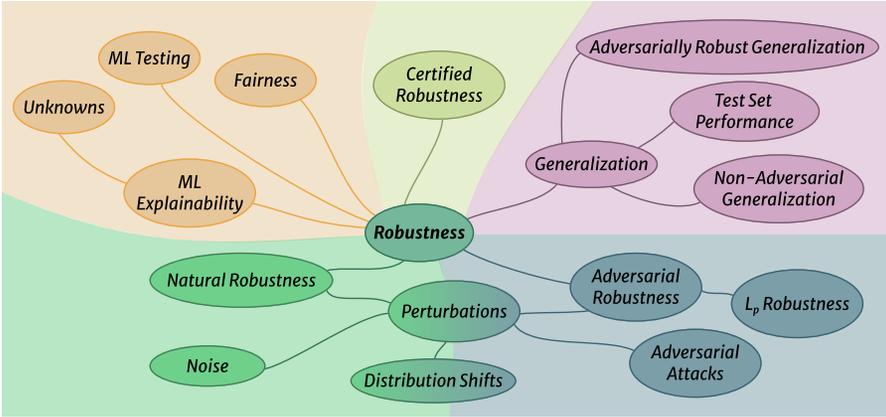

Fig. 2. Main concepts found through our analysis of the literature on Robust AI.

### 3.1 The Various Shades of Robustness

Given the broadness of the literature on robustness and the variety of contexts in which it is considered, addressed, and analyzed, we discuss and provide a common ground about the definitions of robustness and its associated concepts. Particularly, robustness is generally defined as *the insensitivity of a model's performance to miscalculations of its parameters* [155, 268], with Nobandegani et al. [155] stating that *robust models should be insensitive to inaccuracies of their parameters, with little or no decline in their performance*. Two main robustness branches have been identified: robustness to adversarial attacks or perturbations, and robustness to natural perturbations.

*3.1.1 Adversarial Robustness.* Adversarial Robustness refers to the ability of models to maintain their performance under potential adversarial attacks and perturbations [278]. Adversarial perturbations are imperceptible, non-random modifications of the input to change a model's prediction, maximizing its error [217]. The result of such a process is called an adversarial example, i.e., an input $x'$ close to a valid input $x$ according to some distance metric (i.e., similarity), whose outputs are different [38]. Such data is employed to perform adversarial attacks, whose objective is to find any $x'$ according to a given maximum attack distance [45]. The literature presents different classifications of adversarial attacks: targeted and untargeted [44], and white-, grey-, or black-box [147]. Targeted attacks generate adversarial examples misclassified as specific classes, while untargeted attacks generate misclassified samples in general. The main difference between white-, grey-, and black-box attacks is the attacker's knowledge about the model or the defense mechanism.

A similarity metric is often defined when generating attacks or evaluating robustness. Depending on the input domain, different metrics can be applied. These metrics are built as a function of a parameter (usually denoted with the letter $p$) whose value influences its computation. For example, Carlini et al. [38] define a generic $p$ norm from which different metrics with different meanings are derived. In their case, when $p = 0$ ($L_0$ distance), the number of coordinates for which the valid and perturbed input are different is measured; when $p = 2$ ($L_2$ distance), the standard Euclidean distance between the valid and perturbed input is computed; when $p$ = infinite ($L_\infty$ distance), the maximum change to any coordinate is measured. A particular type of robustness is Certified Robustness that guarantees a stable classification for any input within a certain range [53].

*3.1.2 Natural Robustness.* Natural Robustness (a.k.a. Robustness against natural perturbations) is the capability of a model to preserve its performance under naturally-induced image corruptions or alterations. [64]. Natural Perturbations (a.k.a. Common Corruptions [88] or Degradations [78]) are introduced through different types of commonly witnessed natural noise [238], e.g., Gaussian noise in low lighting conditions [88], and represent conditions more likely to occur in the real world compared to adversarial perturbations [64]. Temporal Perturbations are natural perturbations that hinder the capability of a model to detect objects in perceptually similar, nearby frames in videos [191]. All these perturbations result in a condition where the distribution of the test set differs from the one of the training set [108]. This condition is typically referred to in the literature with overlapping concepts, namely distribution shift [60, 220], Out-of-Distribution data (OOD) [79, 196], and data outside the training set [165].

*3.1.3 Generalisation.* Generalisation is another widely used term in the robustness literature. In general, it is defined as the model's performance on unseen test scenarios [163] or as the closeness between the population (or test error) to the training error, even when minimising the training error [153]. Two other types of generalization are also reported: adversarially robust [267] and non-adversarial generalizations [79, 165, 246, 277]. While the first one refers to the capability of a model to achieve high performance on novel adversarial samples, the second one is evaluated on non-adversarial samples (e.g., natural perturbations [246, 277], distribution shifts [79, 165], etc.).

*3.1.4 Performance.* Across the inspected literature, the term performance is employed with a broad variety of meanings. Depending on the aspect of interest, it may refer to accuracy [64], robustness [115], runtime [199], or precision [258]. Given such variety, the actual meaning of performance will be addressed only when relevant to understand the concepts explained in the core survey.

## 3.2 Domains Adjacent to Robustness

Machine learning (ML) explainability, fairness, trustworthiness, and testing, are four research domains recurring across robustness literature. While there is no agreed upon definition of each of these fields and their goals, and we acknowledge it is not possible and desirable in the scope of this survey to provide a complete overview of these fields, we provide here explanations that are sufficient to understand the relation these fields bear to robustness.

*3.2.1 Explainability.* ML explainability is the field interested in developing post-hoc (explainability) methods and (inherently explainable) models that allow the internal functioning of ML systems to be understandable to humans [39]. We identify three types of relations between the explainability and robustness fields. A number of papers investigates how explainability methods can be used in order to *enhance the robustness* of models (see subsubsection 4.2.3). Another set of papers investigates *how robust existing explainability methods are* to various types of perturbations (see subsubsection 5.3.1). A last set of papers instead studies how existing methods for enhancing robustness *trade off* with the explainability of the models, and especially with the alignment between the model features, and the features a human would expect the model to learn (see subsubsection 6.3.3).

We also consider the field of *(un)known unknowns* [136] close to robustness, as they are typically caused by OOD samples. In this field, methods to identify and mitigate the presence of such unknowns are developed and, while these methods typically fall within explainability [193, 229], they are directly applicable to increase the robustness of a model.

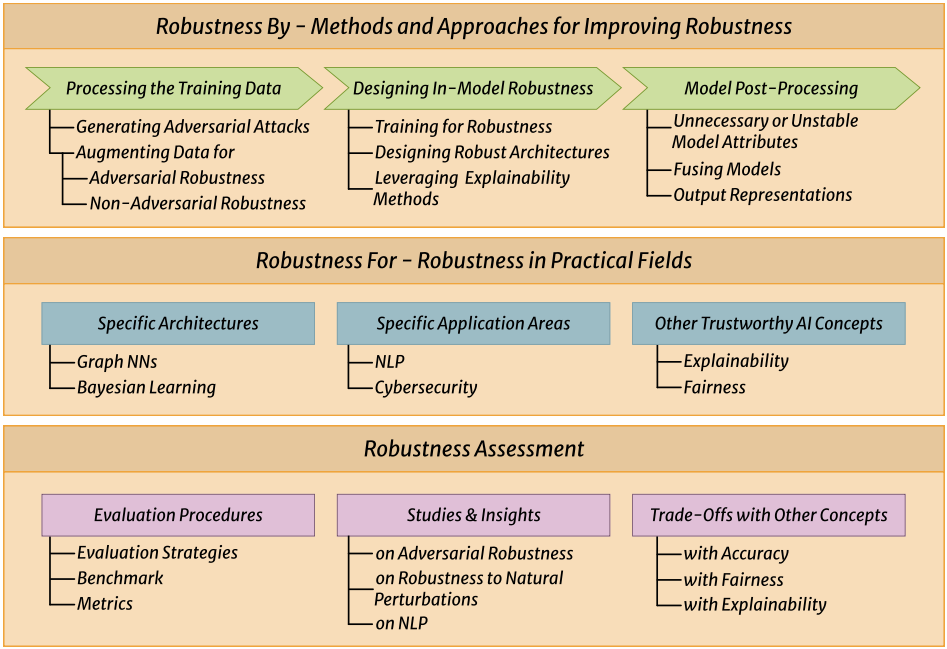

Fig. 3. The three themes and their sub-categories that shape our survey.

*3.2.2 Fairness.* ML fairness in the broad sense is the field interested in making the outputs of an ML model non-harmful to the humans who are subject to the decisions made based on these outputs. Researchers in this field have developed a number of fairness metrics [232] and methods for mitigating unfairness [138]. We identify two types of relations between this field and robustness, similar to the relations between explainability and robustness: *robustness of fairness metrics and methods* to different types of natural and adversarial perturbations (see subsubsection 5.3.2) and *trade-offs* caused by the application of robustness methods (see subsubsection 6.3.2).

*3.2.3 Testing.* ML testing [270] is a field emanated from software testing. It consists in developing methods and tools to identify and characterize any discrepancy between the expected and actual behavior of a ML model. While this field bears a broader scope since brittleness to different perturbations represents one of the many types of unexpected behavior of a model, it is also narrow as it is solely interested in detecting the issue, but not its mitigation. Naturally, methods developed in this field could potentially be adapted in the future to better detect robustness-related issues.

## 3.3 Themes in Relation to These Robustness Shades and Related Domains

Analyzing the collected publications through a thematic analysis approach [30], we iteratively and collaboratively identified three primary themes and three recurring categories within each of these themes (nine categories in total) that were deemed worth emphasizing (summarized in Figure 3).

*3.3.1 Robustness Methods.* The most studied methods to achieve robustness are described in section 4. They are categorized according to the stage of the ML pipeline to which they apply, that is either the processing of the training dataset, the model creation stage, or the post-processing of the trained model. Within each of these stages, the approaches vary across publications, and were further clustered into groups based on types of robustness (e.g., adversarial or natural perturbations), and specific ML component (e.g., training procedure or model architecture) they apply to. For each of the groups, we further delve into sub-groups based on the types of transformation applied to the component (e.g., different loss functions or regularizers), and describe the main similarities and differences across transformations, e.g., in terms of technical approach and performance.

*3.3.2 Robustness in Practical Fields.* While a majority of papers concentrate their studies and the evaluation of their robustness methods around computer vision or do not mention a specific field, we also identify a consequent number of papers that bear different focuses. We separated these papers from the ones discussed above, because they present particularities that are worth investigating. We categorize these papers broadly based on their research fields, and discuss them in section 5. Within each of the categories, we describe the most researched sub-types for which we retrieved the most literature. Particularly, we identified focuses relating to specific model types (Graph Neural Networks and Bayesian Learning), specific application areas (Natural Language Processing, and Cybersecurity), and specific concepts within the trustworthy AI domain (explainability and fairness). The latter is particularly interesting because it differs from other works in its objectives. Contrary to all other papers which investigate model performance under perturbations, it instead investigates evolution of the fairness and explanations of a model under the effect of perturbations.

*3.3.3 Robustness Assessment.* The last theme we identified, described in section 6, revolves around the assessment of the robustness of a system. Particularly, the importance of developing procedures (methodologies, benchmarks, and metrics) to evaluate robustness emerged from the papers and these procedures revealed to vary greatly across publications (be it publications whose primary contribution is an evaluation procedure, or a robustness method that requires to be evaluated through a defined procedure). We also identified a set of publications whose primary objective is to perform studies to evaluate existing robustness methods and collect insights to further characterize in which conditions each type of method performs best. Finally, the last recurring theme was trade-offs, as many papers that propose or evaluate robustness methods tackle trade-offs while striving to achieve other objectives, be it model performance or the other trustworthy AI concepts identified earlier. The publications in this section of the survey are typically falling under the umbrella of computer vision publications, or of the different fields highlighted above.

# 4 ROBUSTNESS BY: METHODS AND APPROACHES FOR IMPROVING ROBUSTNESS

## 4.1 Processing the Training Data

With the final aim of improving model robustness against adversarial attacks, noise, or common perturbations, several approaches focus on generating perturbations to perform data augmentation.

*4.1.1 Generating Adversarial Attacks.* A number of papers tackle the challenge of developing methods to generate adversarial attacks that prove deep learning models brittle. The proposed methods vary with regard to three main objectives. a) The type of task targeted: e.g., natural language processing model [42, 101], image classification [38] or object detection models [46]. b) The type of constraints imposed on the attack: e.g., attack on the physical space before capturing the digital data sample (e.g., by sticking images patches on the physical object to be recognized [238]), or by processing this digital input sample [46]; general attack or attack that targets a particular component of the model (e.g., rationalizers of rationale models [42]); attacks that preserve certain properties of the input sample such as human consistency (e.g., Jin et al. [101] talk about human prediction consistency, semantic similarity, and text fluency with regard to the generated adversarial text samples), additionally to satisfy the constraint on similarity to the original sample [101]. c)

The type of brittleness targeted, e.g., the model makes a different (wrong) prediction when the transformed sample is inputted, or the explanations of the prediction also becomes flawed (i.e., the identified important features are not the correct ones) [247]. The works then differ by the approach taken to generate the attacks, be it through different optimization instances (objective functions) they use to find adversarial instances that fit the problem [38, 46], by leveraging Generative Adversarial Networks (GAN) [1], or through a rule-based algorithmic approach [42, 101].

*4.1.2 Augmenting Data for Adversarial Robustness.* Most of the identified literature focuses on transforming [25, 41, 222, 280], generating [4, 47, 115, 214, 237] or employing ready-to-use [57] data and/or adversarial samples to extend or create datasets to train more robust models. Such a data augmentation process can successfully improve adversarial robustness [41, 57, 115, 214, 222, 237, 280] and adversarial accuracy [4], while sometimes reducing time costs [41], and adversarial attack success rate [25]. When defending against adversarial attacks, GAN-based solutions are proven useful in achieving such an objective [1, 73, 214, 237]. In particular, they are employed to generate adversarial samples [1], perturbations [237], and boundary samples [214] to defend the networks against adversarial attacks. While most methods apply complex transformations to improve robustness, simple transformations, like rotation [222] and image background removal [228], are still proven effective. However, extending the training set is not always enough by itself. Hence, ad-hoc training procedures [41, 47] must be set in place to select [47] and adapt [41] the optimal training data to achieve adversarial robustness.

*4.1.3 Augmenting Data for Non-Adversarial Robustness.* Not all researchers aim to enhance models' defense against adversarial robustness. Noise [171], non-adversarial perturbations [73, 115, 154, 280], spurious correlations [41, 242], and distribution shift [165, 272] hinder the performance and resilience of models. In tackling such impairments, human rationale collection allows the generation of new datasets [165], counterfactual-augmented data [41] and the definition of proper perturbation levels [154], consequently improving performance [41], and model [154], and distributional shift [165] robustness. Custom [171] and pre-existing approaches are applied to perform data augmentation, consequently improving noise robustness [171] and performance [171]. On the other hand, data transformation [73, 280] and training [115] approaches are applied to improve model robustness [73, 115, 280] and reduce training time [73].

## 4.2 Designing In-Model Robustness Strategies

*4.2.1 Training for Robustness.* Training plays an integral part in creating machine learning models. Concerning robustness, Adversarial Training is the de-facto standard for building robust models. The core intuition behind it is to complement natural data with perturbed one such that models incorporate information about data that better represent real-world scenarios' variability. In this section, we discuss adversarial training approaches that adaptively change the perturbation magnitude, allow for the learning of robust features, or include novel loss or regularisation functions. Finally, we discuss approaches alternative to adversarial training.

*Training with Dynamic Perturbations.* In this category, Madaan et al. [132] and Cheng et al. [50] propose methods to generate dynamic perturbations at the level of single data instances that are then controlled by enforcing label consistency in the former case, and smooth labels in the latter. Differently, Rusak et al. [183] devise a neural network-based adversarial noise generator to tackle the online generation of perturbations.

*Training Robust Feature Representations.* An alternative strategy to dynamically enrich data while training is that of pushing the model to learn more robust feature representations. Scholars have achieved this in multiple ways, from designing novel methods altogether [109] to employing additional classifiers [14]. For example, Yang et al. [259] propose to apply perturbations on textual embeddings such that the corresponding words would be drawn toward positive samples rather than adversarial ones. Bai et al. [14] take a modeling approach to obtain robust features through the addition of auxiliary models to identify which channels in CNNs are more robust.

*Adversarial Training Algorithms.* Adversarial Training has proven to be a fundamental tool to build robust models and that is reflected in the amount of literature available for it: researchers have focused in improving the whole process and proposed a plethora of algorithms [85, 133, 212, 221, 241], borrowing different ML paradigms like self-supervised and unsupervised learning [152, 226], that are applicable to a variety of tasks, e.g., content recommendation [248, 264]. In this context, Projected Gradient Descent (PGD) [133] is a common white-box (i.e., the attacker knows everything about the model) algorithm. On the same note, Terzi et al. [221] and Gupta et al. [85] propose extensions of PGD by using Wasserstein distance in the adversarial search space, by replacing the initial adversarial training stages with *natural* training, or by encouraging the logits from clean examples to be similar to their adversarial counterparts, respectively. On the other hand, several works focus on leveraging other types of information. For example, Zoran et al. [284] adversarially train and analyze a neural model incorporating a human-inspired, visual attention component guided by a recurrent top-down sequential process. Shifting to model outputs, works from Wang et al. [241] and Stutz et al. [212] focus on differently treating misclassifications and rejecting low-confidence predictions. Similarly, Haase-Schütz et al. [86] and Cheng et al. [49] deal with progressively tuning labels starting from unlabelled data and through smoothing, respectively.

*Training with Adapted Regularizers.* Regularisation is another tool that ML engineers can use when building models and, as such, it has also been used to make them more robust. Li and Zhang [120] propose a PAC-Bayesian approach to tackle the memorization of training labels in fine-tuning. Chan et al. [40] suggest an approach that optimizes the saliency of classifiers' Jacobian by adversarially regularizing the model's Jacobian to resemble natural training images.

*Training with Adapted Loss Functions.* Loss functions are essential objectives used to train ML models. Concerning robustness, a variety of loss functions have been used to incorporate specific objectives: triplet loss [135], minimising distance between true and false classes [123], mutual information [253], consistency across data augmentation strategies [218], perturbation regularizers [255], adding maximal class separation constraints [149], combining multiple losses [107] (e.g., Softmax and Center Loss), or approximating existing losses (e.g., Categorical Crossentropy) [68]. It is worth noting that loss functions tailored for robustness are not exclusive to models trained in isolation and robust and natural models (acting as regularizers) can be jointly trained [9].

*Beyond Adversarially Training.* Researchers have studied alternative training procedures to adversarial training. Staib [210] has analyzed the relationship between adversarial training and robust optimization, proposing a generalization of the former which leads to stronger adversaries. Attention is also directed to leveraging input and output spaces. Li et al. [122] consider training robust models by leveraging the adversarial space of another model. Differently, Mirman et al. [143] and Rozsa et al. [181] leverage abstract interpretation and evolution stalling, respectively. The former generates abstract transformers to train certifiably robust models. The latter progressively tempers the contributions of correct predictions toward the loss function. Finally, Mirman et al. [143], Zi et al. [282], and Papernot et al. [162] leverage Distillation (a knowledge transfer technique in which a smaller model is trained to mimic a larger one) [31, 90] to obtain robust models.

### 4.2.2 Designing Robust Architectures.
In the context of Robust AI, researchers have also investigated possible ways to make neural network models robust from an architectural perspective.

*Tweaking Neural Network Layers.* We identified that a considerable amount of effort is directed toward Computer Vision applications, with many solutions aimed at integrating additional mechanisms of Convolutional Networks to enhance their robustness. Many adversarial attackers create harmful data instances by injecting noise perturbations in the input of the model. In line with this, many researchers have attempted to introduce mechanisms that take advantage of this information or directly try to mitigate the repercussions of such perturbations. For example, Jin et al. [102] introduce additive stochastic noise in the input layer of a CNN and re-parametrize the subsequent layers to take advantage of this additional information. Alternatively, Momeny et al. [146] introduce a CNN variant that is robust to noise by adapting dynamically both striding of convolutions and the following pooling operations. Work by Xu et al. [251] operate on the classification layer by constraining its weights to be orthogonal. Operating on network layers is not exclusive to the aforementioned discriminative models, but it has also found applications for generative models. For example, Kaneko et al. [104] propose a method to obtain Generative Adversarial Networks (GAN) that do not require a large amount of correctly-labeled instances but still maintain a consistent behavior. They do this by integrating a noise transition model that maps clean and noisy labels which leads to GANs that are resilient to different magnitudes of label noise.

*Leveraging the Inherent Robustness of Spiking Neural Networks (SNN).* In parallel to such enhancements at the architectural level, a growing trend is represented by SNN [131]. SNNs are a particular type of neural network that mimics the behavior of biological neurons by incorporating the notion of time and both operating with and producing sequences of discrete events (i.e., spikes). Concretely, a neuron in a SNN transmits information only when its value surpasses a certain threshold. This particular kind of neural network was found to be inherently robust to certain types of adversarial attacks. Sharmin et al. [194] test Spiking Neural Networks directly against gradient-based (black-box) attacks and find that such architectures perform better than non-spiking counterparts without any kind of adversarial training. Inspired by neuroscience, Cheng et al. [51] formulate Lateral Interactions (i.e., intra-layer connections) for SNNs which provide both better efficiency when processing a series of spikes as well as better resistance to injected Gaussian noise.

*Searching Neural Architectures.* Connected to handcrafting robust neural architectures, scholars have started applying Neural Architecture Seach (NAS) to such a problem. In general, NAS is an automatic procedure aimed at discovering the best architecture (e.g., in terms of accuracy) for a neural network for a specific task. Devaguptapu et al. [58] analyze the effects that a varying amount of parameters have on adversarial robustness: while NAS can be an alternative to adversarial training, handcrafted models are more robust on large datasets and against stronger attacks like PGD [133]. Their insights motivate other works in this space, that focus on strengthening NAS approaches by including different forms of regularization on the smoothness of the loss landscape [145], or the sensitivity of the network [62, 94]. A different take on using NAS is the one of Li et al. [124]: architecture search was blended with existing models (e.g., ResNet) to find the minimal increase in model capacity allowing it to withstand adversarial attacks.

*The Case of Non-Neural Models.* Despite increased interest in neural networks, the robustness of other machine learning models is still an open problem being investigated. To this avail, Chen et al. [43] frame the task of learning Decision Trees as a max-min saddle point problem which, by approximating the minimizer in the saddle point problem, lead to Decision Trees that proved to be robust to adversarial attacks. On this note, Singla et al. [200] show the benefits of applying Correntropy [129] to the $\alpha$-hinge loss function used to train Support Vector Machines, which results in models achieving robustness to label noise while maintaining competitive performance.

*4.2.3 Leveraging Explainability Methods.* In turn, using explainability to make robustness easier to understand has also received some attention. A few works have investigated to what extent existing explainability methods can be adapted in order to increase model robustness. Especially, Kortylewski et al. [109] propose *Compositional Neural Networks*, a unification of convolutional neural networks (CNN) with part-based models (inherently interpretable models), and show that these new networks increase model robustness to various partial occlusions of objects. Chen et al. [42] also demonstrated that inherently interpretable models such as rationale models in NLP are naturally more robust to certain adversarial attacks yet are still brittle to certain scenarios.

## 4.3 Leveraging Model Post-Processing Opportunities

Robustness can also be improved through methodologies applied after training the model.

*4.3.1 Identifying Unnecessary or Unstable Model Attributes (neurons, features).* Pruning, i.e., the act of removing neurons and/or connections from a model, has become a popular compression approach that aims at reducing the computational cost of training models [125]. Recent literature in Robust AI has explored the use of pruning techniques or methodologies inspired by pruning to enhance model robustness [42]. Chen et al. [42], for instance, design a methodology for selectively replacing ReLU neurons that are identified as unstable (i.e., neurons that operate in the flat area of the function) and insignificant by linear activation functions that help improve robustness at a minimal performance cost. In a similar vein, additional mechanisms have been suggested for dealing with unnecessary and/or unstable system attributes. For instance, Gao et al. [72] introduce DeepCloak, a novel method to detect and remove unnecessary classification features in deep neural networks, consequently reducing the capabilities of attackers to generate such attacks.

*4.3.2 Fusing Models.*

*Against Input Issues.* Another approach for achieving post-model-training robustness consists of plugging additional models into a trained model. These additional models can be used to identify and deal with problematic data instances (e.g., out-of-distribution, mistaken [170], noisy [180], or adversarially modified [261]). For instance, in the context of Natural Language Processing, Pruthi et al. [170] attach a task-agnostic word recognition model to a classification model as a means to defend the main classifier against spelling mistakes. In the context of Computer Vision, Ye et al. [261] use an additional classifier to determine real vs. adversarially manipulated data instances. This additional classifier would receive an overlap of the data instance and its saliency map.

*Against infected models.* Model fusion is also used to identify and deal with infected models (e.g., backdoor-infected neural networks [208]), to compare the robustness of small models with respect to compression techniques [245]. A prominent line of work in this field consists in using Generative Adversarial Networks as auxiliary models. This strategy has been used for dealing with input data [214] and models [52]. For the former, Sun et al. [214] use a Boundary Conditional GAN to generate boundary samples. These boundary samples have true labels and are near the decision boundary of a pre-trained classifier. For the latter, Choi et al. [52] propose Adversarially Robust GAN (ARGAN), that trains the generator model to reflect the vulnerability of the target neural network model against adversarial examples and hence optimizes its parameter values.

*4.3.3 Applying Changes in Output Representations.* A final approach consists of varying the representation of the output. To this end, Verma and Swami [231] suggest an approach for designing an Error Correcting Output Code that moves away from one-hot encoding of outputs to an encoding with a larger Hemming distance (> 2). This forces adversarial perturbations to be larger.

# 5 ROBUSTNESS FOR: ROBUSTNESS IN PRACTICAL FIELDS

## 5.1 Robustness for Specific Architectures

*5.1.1 Attacks on Graph Neural Networks (GNN).* A number of papers investigate how to increase the robustness of specific types of model architectures. One of the most prominent ones is GNN. GNN are susceptible to adversarial attacks due to small adversarial perturbations having a large effect on their output. Hence, there has been research into making such networks more robust to attacks. Pezeshkpour et al. [166] propose an attack strategy that generates adversarial examples, which find the minimal changes necessary to make in the link prediction task that causes a label change, for link prediction problems on knowledge graphs. An approach by Lou et al. [130] determines controllability and connectivity robustness, indicating how well a system can keep its connectedness and controllability against node- or edge-removal attacks, by compressing the high-dimensional adjacency matrix to a low-dimensional representation before feeding it to a Convolutional Neural Network to perform the robustness prediction. Fox and Rajamanickam [71] investigate the impact of structural noise on the robustness of GNN and find them to be weak to both local and global structural noise. Geisler et al. [76] focus on particularly large graphs and propose new attack and defense strategies for this case to improve the efficacy of attacks on GNN and a defense mechanism with a low memory footprint that enables defenses on large networks at scale. Significant attention has also been paid to formally certifying the robustness of GNN [28, 236].

*5.1.2 New Frameworks for Graph Neural Networks.* There have also been several proposals for new GNN frameworks that have better robustness characteristics. For example, Jin et al. [103] establish a framework to jointly learn clean graph structures from perturbed ones as well as the parameters of a GNN that is robust to adversarial attacks. They do so by iteratively reconstructing the clean graph by preserving low rank, sparsity, and feature smoothness properties, allowing them to eliminate edges that have been crafted by an adversary. Another end-to-end learning framework [48] was put forward, that jointly and iteratively learns graph structure and graph embeddings, where a similarity metric and adaptive graph regularization are applied to control the quality of the learned graph. Zhang and Lu [271] introduce a model where robustness to noise is achieved by a shared auxiliary for neighborhood aggregation, using a new aggregator function making use of masks. Here, the auxiliary model learns a mask for each neighbor of a given node, making node-level and feature-level attention possible. Thus, it is capable of assigning different importance values to both nodes and features for predictions, which increases robustness.

*5.1.3 Bayesian Learning.* Many adversarial attack strategies are based on identifying directions of high variability and since such variability can be intuitively linked to uncertainty in the prediction, Bayesian Neural Networks are naturally of interest for robustness research. Pang et al. [161] evaluate the robustness of Bayesian networks against adversarial attacks for image classification tasks. Similarly, Carbone et al. [36] analyze Bayesian networks to show that they are robust to gradient-based attacks. Vadera et al. [227] focus on different inference methods and attacks whose goal is to cause the model to misclassify the provided input to evaluate the network robustness. They find that Markov Chain Monte Carlo inference has excellent robustness to different attacks. Miller et al. [140] aim to evaluate robustness by extracting label uncertainty from the object detection system via dropout sampling. They perform uncertainty estimation through dropout sampling to approximate Bayesian inference over the parameters of deep neural networks and find that the estimated label uncertainty can be used to increase object detection performance under open-set conditions.

## 5.2 Robustness for Specific Application Areas

*5.2.1 Robustness for Natural Language Processing (NLP) Tasks.* The robustness of NLP systems is paramount. Adversarial attacks and training both represent active areas of research in recent years. They aim to make NLP models less susceptible to attacks and noisy data, therefore improving robustness. As such, a multitude of approaches have been proposed specifically for this domain.

Zheng et al. [279] present an approach to study both where and how parsers make mistakes by searching over perturbations to existing texts at the sentence and phrase levels. Furthermore, they design algorithms to create such examples for white-box and black-box models. They demonstrate that parsing models are susceptible to adversarial attacks. At the word level, Yang et al. [259] propose a method designed to tackle word-level adversarial attacks by pulling words closer to their positive samples while pushing away negative samples. They find that their method improves model robustness against a wide set of adversarial attacks while keeping classification accuracy constant. Similarly, Du et al. [66] study the weakness of many state-of-the-art NLP models against word-level adversarial attacks and propose Robust Adversarial Training to improve the models' robustness against adversarial attacks. Pruthi et al. [170] look to combat adversarial misspellings by attaching a word recognition model to the classification model. They find that the adversary can degrade the performance of a text classifier to the point where it is equivalent to random guessing just by altering two characters per sentence. Zhou et al. [281] employ multi-task learning, where a transformer-based translation model is augmented with two decoders with different learning objectives, to improve the robustness to noisy text. Similarly, Li et al. [121] use adversarial multi-modal embeddings and neural machine translation to denoise input text, making it effective against adversarially obfuscated texts. Chen et al. [42] inspect whether NLP models are capable of generating rationales (i.e., subjects of inputs that can explain their prediction) to provide robustness to adversarial attacks. They find that rationale models show promise in providing robustness, though their robustness is highly variable.

*5.2.2 Robustness for Cybersecurity.* Cybersecurity deals with the resistance to intelligent attacks, as such, there has been research into the robustness of cybersecurity systems, in particular into the robustness of AI used in cybersecurity applications. A significant focus has been the robustness of malware detection. For instance, Abusnaina et al. [2] improve malware classifier accuracy by using Control Flow Graphs extracted from the attacked code, which represent behavior patterns, as the input data for their threat detector. These are subsequently altered to obtain adversarial examples and test the robustness of the overall model.

There has also been research into the security, and specifically malware detection, of specific operating systems or platforms. Anupama et al. [8] initially use the Fisher score to identify and select the most relevant attributes for the classifiers and subsequently develop three different attack approaches to create adversarial examples. The evaluation of the resulting classifiers finds that this approach greatly increases detection rates.

Beyond this, defense against distributed denial-of-service (DDoS) attacks, where a service or network is overwhelmed with additional artificial traffic, has been studied through the lens of robustness as well. Abdelaty et al. [1] present an adversarial, GAN-based training framework to produce strong adversarial examples for the DDoS domain to tackle the weaknesses of Network Intrusion Detection Systems against adversarial attacks. The generator model of a GAN, trained on benign samples, is used to produce adversarial samples. Then, DDoS samples are perturbed by changing their features using values taken from the generated examples. The effectiveness of such attacks is thus greatly decreased. Amarasinghe et al. [6] apply Layer-wise Relevance Propagation to the trained anomaly detector, yielding relevance scores for each individual feature.

## 5.3 Robustness for Other Trustworthy AI Concepts

*5.3.1 Robustness for Explainability.* Robustness has been widely discussed in the context of explainability methods in recent years [61, 114]. For explainability to be implemented effectively in ML systems, it must be robust. More robust explanations will naturally lead to more trustworthy explainable AI, as humans can feel more secure regarding phenomena such as adversarial attacks on the system and out-of-distribution input data. A critical step toward more robust explainability is the effort to provide methods for evaluating explanations with respect to robustness.

A method for measuring explainability was proposed by Zhang et al. [273], whose approach explores the input space to measure the percentage of inputs on which the prediction can be consistently explained with the simple model height of the decision tree used to explain the neural network's prediction. However, their result is inconclusive as it may seem tied to imbalances in the data used. In a similar vein, Nanda et al. [150] propose a scalable framework using machine-checkable concepts to assess the quality of generated explanations with respect to robustness, specifically their vulnerability to adversarial attacks. Alvarez-Melis et al. [5] define a novel notion of robustness based on the point-wise, neighborhood-based local Lipschitz continuity. Gradient- and perturbation-based interpretability methods are evaluated, revealing the non-robustness of such practices and the high instability of perturbation-based methods.

Atmakuri et al. [13] focus on understanding the adversarial robustness of explanation methods in the context of text modality. In particular, they utilize saliency maps to generate adversarial examples to evaluate the robustness of the model of interest. They find the used Integrated Gradient explanation method is weak against misspelling and synonym substitution attacks.

*Robustness for Counterfactual Explanations.* There have also been multiple works on the robustness of generated counterfactual explanations to adversarial inputs. Virgolin and Fracaros [233] explore how to improve such robustness by giving a formal definition of what it means to be robust towards perturbations and implementing this definition into a loss function. To test this definition, they release five datasets in the area of fair ML with reasonable perturbations and plausibility constraints. They find that robust counterfactuals can be found systematically if we account for robustness in the search process. Further, Pawelczyk et al. [164] explore counterfactual explanations by formalizing the similarities between popular counterfactual explanations and adversarial example generation methods, identifying conditions when they are equivalent. Thus, they derive the upper bounds on the distances between the solutions output by counterfactual explanation and adversarial example generation methods. Bajaj et al. [15] generate robust counterfactual explanations on GNNs by explicitly modeling the common decision logic of GNNs on similar input graphs. The generated explanations are naturally robust to noise because they are produced from the common decision boundaries of a GNN that govern the predictions of many similar input graphs. The generation of robust text-based counterfactual explanations has also been studied for NLP tasks [242, 263]. Finally, several works focus on making the popular interpretable model-agnostic explanations (LIME) approach more robust to adversarial attacks [184, 204].

*5.3.2 Robustness for Fairness.* A key attribute of any system to be put into production is fairness. The relationship between fairness and robustness, and how one contributes to the other, has received increased attention recently. Rezaei et al. [175] aim to make classifications that have robust fairness without relying on previously labeled data, as these may carry some inherent biases. Wang et al. [240] study the effect of relying on noisy protected group labels, providing a bound on the fairness violation concerning the true group. Yurochkin et al. [265] propose an adversarial approach to fairness, using a distributionally robust approach to enforcing individual fairness during training.

There have also been efforts to improve the fairness of graph-based counterfactual explanations, such as Agarwal et al. [3] who aim to establish a connection between counterfactual fairness and graph stability by developing layer-wise weight normalization and therefore enforcing fairness and stability in the objective function. They see increases in fairness and stability without a decrease in performance. Further, Bajaj et al. [15] propose a method to generate robust counterfactual explanations on GNN by explicitly modeling the common decision logic on similar input graphs. The explanations are naturally robust to noise because they are produced from the common decision boundaries of a GNN that govern the predictions of many similar input graphs.

# 6 ROBUSTNESS ASSESSMENT AND INSIGHTS

## 6.1 Evaluation Procedures

*6.1.1 Evaluation Strategies.* A fundamental aspect of interest associated with robustness is its evaluation, i.e., applying techniques, benchmarks, and metrics to assess its degree empirically. Given the multifaceted nature of robustness, a wide variety of approaches have been developed. In particular, two different branches can be identified, based on whether robustness is certified or not.

*Evaluation of Robustness.* Concerning the first group, most methodologies either compute a safe radius [113, 182] or region [82] within which the model performs robustly, or they compute their complementary region [275], i.e., error regions. Abstract Interpretation, i.e., *a theory which dictates how to obtain sound, computable, and precise finite approximations of potentially infinite sets of behaviors* [75], enables robustness evaluation when combined with techniques like constraint solving [258] and importance sampling [134]. Other evaluation approaches reformulate the robustness assessment problem from different perspectives. Tjeng et al. [223] formulate the verification of the robustness against adversarial attacks as a mixed integer linear program by expressing properties like adversarial accuracy as a conjunction, or disjunction, of linear properties over some set of polyhedra. Webb et al. [243] statistically evaluate robustness by estimating the proportion of inputs for which a defined adversarial property (i.e., an adversarial condition associated to a function that evaluates its violation) is unsatisfied (i.e., there are no counterexamples violating such a property). This reframing is useful to widen the variety of solutions that can be applied to assess robustness, consequently improving their scalability [223, 243], computational speed [223, 260], and enabling the application of pre-existing tools [89].

*Evaluation of Certified Robustness.* When it comes to assessing certified robustness, most of the literature focuses on model robustness against adversarial attacks [65, 99, 118, 119, 199, 201, 202, 276]. To this end, researchers focus on the efficient computation of robustness bounds [65, 118, 201, 276] while also improving the training procedure to achieve efficiently certifiable [276],or ready to certify [99], models. Deterministic [119] and Random [67] Smoothing approaches have also proven to be effective in evaluating $L_1$ [119] and $L_2$ robustness. Nevertheless, overapproximation [199], orthogonalization relaxation [202], and regularization [99] have also been successfully applied to improve the computation of certifiable bounds in adversarial settings. Moreover, Zhang et al. [269] strive to generalize certification techniques to non-piecewise linear activation functions.

Even though most literature focuses on certifying the robustness of models against adversarial attacks, other types of certified robustness have also been assessed, e.g., certified robustness to random input noise from samples and geometric robustness [19].

*6.1.2 Benchmarks.* Even though most of the literature on assessing robustness is focused on designing methodologies to evaluate model robustness, some works also propose comprehensive benchmarks, i.e., standardized methods including an approach, dataset, and pipeline to evaluate the robustness of a model against a specific set of attacks. In Computer Vision, robustness against various types of adversarial attacks [63, 81, 167] and common corruptions [88, 137], including noise [88, 278], has been evaluated through benchmarking on datasets [88, 137, 167], with custom measures [63, 88], or using comprehensive frameworks [219]. In the first case, pictures are altered through adversarial or common perturbations (e.g., noise, blur, etc.) [88, 137, 219], and either generalizability [137] (i.e., whether the model can adequately classify newly perturbed pictures) or its behavior by means of custom metrics [63, 88] are evaluated. A few benchmarks have also been applied in the context of graph networks and Natural Language Processing. For instance, Zheng et al. [278] develop scalable datasets to standardise the process of attack and defence, covering graph modification and graph injection attacks.

*Target of Benchmarks.* While some benchmarks are focused on evaluating the effectiveness of defence methods [63, 278], others focus on the intrinsic robustness of the architecture [55, 219]. Tang et al. [219] benchmark architecture design and training techniques against adversarial and natural perturbations, and system noise by developing a platform for comprehensive robustness evaluation. It includes pre-trained models to enhance the evaluation process and a view dedicated to understanding mechanisms for designing robust DNNs. While the main objective of a robustness benchmark is to standardise the approach and/or the data to evaluate such a property, it can also be employed to compare and organise evaluation methods and robust models, allowing researchers to reliably access resources [55]. Regarding comparisons of model robustness, most benchmarks use well-known datasets (e.g., MNIST or ImageNet), implicitly assuming the data is always correct. Some authors have argued that such an assumption should not be lightly accepted since potential errors may influence the benchmarking process and its results [156]. Such a premise can be especially problematic when comparing performances of different models as they can be affected differently by such errors [156]. From a broader perspective, evaluating model robustness can be seen as a part of a process to evaluate fairness. Driven by such an objective, Ding et al. [59] create a series of datasets to evaluate different aspects of ML approaches and benchmark their fairness with respect to noise and data distribution shift.

### 6.1.3 Metrics.
When it comes to the evaluation of the robustness of a model, not only is it essential to choose the proper method, but it is also fundamental for the applied measure to represent model robustness properly. To this end, several measures of robustness have been proposed.

*Metrics for Adversarial Robustness.* Most of the collected literature focuses on describing metrics to evaluate the robustness of networks against adversarial attacks [244, 262]. These metrics are generated by converting the robustness analysis into a local Lipschitz constant estimation problem [244], or qualitatively interpreting the adversarial attack and defence mechanism through loss visualisation [262]. Moreover, only a few of them [244] aim to disentangle the relationships between the evaluation process and the model or attack employed, consequently building model-agnostic [244] and attack-agnostic metrics.

*Metrics for Adversarial Attacks.* Other researchers focus on suggesting metrics for different aspects of adversarial attacks, devising approaches for evaluating the convergence stability of adversarial examples generation [116] and comparing adversarial attack algorithms [23]. Beyond the necessity for metrics to assess model robustness, other metrics have been proven useful in evaluating aspects of robustness [10, 159], like its relationship with adversarial examples [10] and accuracy [159].

*Metrics Outside of Computer Vision.* While most of the literature addresses robustness in Computer Vision, a small part of the literature discusses robustness in other contexts. In the context of NLP, extending the concept of robustness through a metric aligned with linguistic fidelity has been proven effective in improving performances on complex linguistic phenomena [112]. However, recent research [35] have shed light on the lack of robustness of such metrics in the context of tree-based classifiers. Consequently, an extension of robustness through a resilience measure that considers both robustness and stability has been proposed. Such findings highlight the need for creating sound and robust metrics while also addressing less covered contexts.

*Metrics for the Complexity of Robustness Methods.* While most literature focuses on implementing evaluation approaches, a small research branch focuses on improving their efficiency, mainly enhancing their precision in computing the robustness bounds [199, 258] and reducing their computational complexity [215, 235] and execution time [249].

## 6.2 Studies around Proposed Robustness Methods & Insights

*6.2.1 Insights on Adversarial Robustness.* Studying the adversarial robustness of different machine learning techniques has been a persistent research focus in recent years.

*Based on Comparisons.* Beyond formal methods and frameworks, there are several examples of papers empirically evaluating robustness through comparison [100, 189]. For instance, Jere et al. [100] compared the generalization capabilities of convolutional neural networks and their eigenvalues and further compared what features are exploited by naturally trained and adversarially trained models. They found that for the same dataset, naturally trained models exploit high-level human-imperceptible features and adversarially robust models exploit low-level human-perceptible features. Another example in this line is the work by Sehwag et al. [189] who inspected the transferability of the robustness of classifiers trained on proxy distributions from generative model to real data distribution, discovering that the difference between the robustness of classifiers trained on such datasets is upper bounded by the Wasserstein distance between them.

*Based on the Investigation of Activation Function and Weights Perturbations.* There have been several works studying the robustness of models under perturbation of weights or due to changes in activation functions. For example, Tsai et al. [224] studied the robustness of feed-forward neural networks in a pairwise class margin and their generalization behavior under different types of weight perturbation. Furthermore, they designed a novel loss function for training generalizable and robust neural networks against weight perturbation. Song et al. [207] showed that adversarial training is not directly applicable to quantized networks. They proposed a solution to minimize adversarial and quantization losses with better resistance to white- and black-box attacks. Another work that focused on such attacks is Shao et al. [192], who studied the robustness of vision transformers against adversarial perturbations under various black-box and white-box settings.

*6.2.2 Insights on Robustness to Natural Perturbations.* There has also been substantial research into model robustness to noise and out-of-distribution data.

*About Robustness to Noise.* A prominent line of work is evaluating robustness of AI systems against noise [21, 283]. Some examples in this area include the study conducted by Ziyadinov and Tereshonok [283], who evaluated whether training convolutional neural networks using noisy data increases their generalization capabilities and resilience against adversarial attacks. They found that the amount of uncertainty in the training dataset affects both the recognition accuracy and the dependence of the recognition accuracy on the uncertainty in the testing dataset. Furthermore, they showed that a dataset with such uncertainty can improve recognition accuracy, consequently enhancing its generalizability and resilience against adversarial attacks. Bar et al. [21] also evaluated the robustness of deep neural networks to label noise by applying spectral analysis. The authors demonstrated that regularizing the network Jacobian reduces the high frequency in the learned mapping and show the effectiveness of Spectral Normalization in increasing the robustness of the network, independently from the architecture and the dataset.

*About Robustness to Differences in Distributions.* Another area of interest is studying differences in data distributions. On the problem of object-centric learning, Dittadi et al. [60] discovered that the overall segmentation performance and downstream prediction of in-distribution objects is not affected by a single out-of-distribution object. On the other hand, Burns and Steinhardt [33] studied adaptive batch normalization, which aligns mean and variance of each channel in CNNs across two distributions . They found that for distribution shifts that do not involve changes in local image statistics, accuracy can be degraded because of batch normalization.

*6.2.3   Insights for the Natural Language Processing (NLP) Context.* In order to evaluate robustness in NLP, Moradi and Samwald [148] designed and implemented character- and word-level perturbations to simulate scenarios in which input text is noisy, or different from training data. Such perturbations were employed to evaluate the robustness of different language models to noisy inputs. They found that the inspected language models were susceptible to the proposed perturbations, small ones as well. Instead, La Malfa et al. [113] proposed an concept of *semantic robustness*. It generalizes the notion of robustness in NLP by explicitly measuring cogent linguistic phenomena, aligning with human linguistic fidelity, while being characterized in terms of the biases it may introduce. On the other hand, Wang et al. [239] proposed a dataset for evaluating trustworthiness and robustness. It collects annotations, and respective human explanations, that cover different types of adversarial attacks (e.g., obscure expression) and allow for multiple sentiment labels for a single text sample.

Specific architectures, such as Natural Language Inference models have also been evaluated. Sanchez et al. [186] found that these models suffer from insensitivity to small but semantically significant alterations while also being influenced by simple statistical correlations between words and training labels. In addition, the models were insensitive to the proposed transformation on the input data and they exploit a bias polarity to stay robust when new instances are shown.

## 6.3   Trade-Offs Between Robustness and Other Trustworthy AI Objectives

*6.3.1   Trade-Off Between Robustness and Accuracy.* A key question to be asked when analyzing the robustness of a system is what the impact of the changes is on the accuracy of the model. Multiple studies have found a significant trade-off between robustness and accuracy, where an increase in one leads to a decrease in the other. Su et al. [213] evaluated the robustness of 18 existing deep image classification models, focusing on the trade-off between robustness and performance. They found that model architecture is a more critical factor to robustness than model size and that networks of the same family share similar robustness properties. Raghunathan et al. [173] further discussed this and described in detail the effect of augmentation achieved through adversarial training on the standard error in linear regression models when the predictor has zero standard and robust error. Tsipras et al. [225] also studied how robustness and accuracy trade-off, as well as the features that were learned. While Miller et al. [142] investigated the connection between accuracy in- and out-of-distribution and show that that out-of-distribution performance is strongly correlated with in-distribution performance for a wide range of models and distribution shifts.

*6.3.2   Trade-Off Between Robustness and Fairness.* Benz et al. [24] evaluated the impact of robustness on accuracy and fairness. They found inter-class discrepancies in accuracy and robustness, specifically in adversarially trained models and that adaptively adjusting class-wise loss weights negatively affects overall performance. Xu et al. [252] hypothesized that adversarial training algorithms tend to introduce severe disparity in accuracy and robustness between different groups of data, and showed this phenomenon can happen under adversarial training algorithms minimizing neural network models' robustness errors. They also propose a Fair-Robust-Learning framework to mitigate unfairness in adversarial defenses. On the other hand, Pruksachatkun et al. [169] studied if an increase in robustness can improve fairness. They investigated the utility of certified word substitution robustness methods to improve the *equality of odds* and *equality of opportunity* in text classification tasks. They found that certified robustness methods improve fairness, and using both robustness and bias mitigation methods in training results in an improvement for both.

*6.3.3 Trade-Off Between Robustness and Explainability.* Few works investigate the extent to which methods for increasing model robustness impact the features such models use to make predictions, and especially to what extent these features remain meaningful to human judgement. Especially, Woods et al. [247] showed that the fidelity of explanations is negatively impacted by adversarial attacks, and propose a regularisation method for increasing robustness lead to better model explanations (termed *Adversarial Explanations*). Nourelahi et al. [157] investigated how methods dealing with out-of-distribution examples impact the alignment of the features the model has learned with features a human would expect to use. While this is an initial empirical exploration, their results illustrate the complexity of the relation between robustness and feature alignment, as there does not seem to be a model that performs consistently better over these criteria. They suggest to extend their benchmark effort to more types of models, and of robustness and explainability techniques.

# 7 DISCUSSION: DISPARATE RESEARCH ON THE VARIOUS FACETS OF ROBUSTNESS

The robustness of AI systems is a broad, open problem under the umbrella of Trustworthy AI and the copious amount of literature that can be found is a testament to that. Researchers from diverse domains have studied the impact of controlled data perturbations as well as naturally-occurring ones, how to strengthen neural architectures through additional mechanisms, and how to efficiently and effectively train models underlying such systems. In this section, we summarize the gaps and trends we evinced from our inspection of the existing literature.

## 7.1 Addressing Gaps from the Literature

### 7.1.1 Gaps within Robustness.

*The Computer Vision Hegemony.* The immediate outcome of our survey is the extensive effort put into studying — and enhancing — the robustness of models targeted toward Computer Vision, especially Convolutional Neural Networks. Papers from that sub-field of Artificial Intelligence greatly outnumber the ones from other areas, like Natural Language Processing. We found this to be the case regardless of the aspect (attack generation, defense, etc.) scholars focus on. While important, such a focus being put on Computer Vision only begs the question of why other domains have received little contributions compared to the former. Possible explanations for this can be traced back to difficulties in defining perturbations and attacks within certain data manifolds, e.g., word embeddings, or to the lack of alignment between robustness in machine learning and robustness in specific application domains, e.g., signal processing. On the other hand, the intrinsic complexity of pictures compared to other types of data, in particular with respect to the features that can be perturbed and the diversity in the available approaches to evaluate distances between pictures, influence the broadness of the research field.

*Natural Brittleness.* Regardless of the domain, we found that little attention is put on defining natural perturbations and attacks. Instead, much work revolves around defining synthetic attacks and evaluating defense mechanisms against them. While this may make sense from the perspective of a malicious attacker, it does not necessarily translate to robustness in real-world operating conditions. Only a few works in Computer Vision focus on such a type of attacks. Another interesting research direction is signaled by the lack of model-agnostic adversaries. While both automatic and rule-based approaches to generating adversaries exist, these tend to be targeted toward certain types of AI systems. Obtaining model-agnostic attacks would be the dual case to such a scenario and could provide for a common baseline for evaluating the robustness of AI systems. Moreover, achieving model-agnostic and perturbation-agnostic evaluations approaches would allow to disentangle the relationship between these scenario-specific aspects and the actual robustness of the model, finally leading to an unbiased analysis of the robustness of a system [244].

*7.1.2   Gaps Stemming from the Intersection Between Robustness and Other Trustworthy AI Concepts.*

*Robustness and Explainability.* Considering the brittleness of existing AI systems in conjunction with their opaqueness, their explainability is of paramount importance. Explainable AI (XAI) methods have been, and still are being, proposed [83, 84] to tackle such a challenge. However, few works discuss the robustness of XAI methods and the produced explanations. This is a crucial dimension that needs to be addressed to obtain explanations that are both faithful (i.e., correctly describing model behavior) and trustworthy. By the same token, explainability can better inform the ideation and implementation of approaches geared towards robustness.

*Tensions between Accuracy, Robustness, Fairness, and Explainability.* Connected to the above points, it is worth noting how existing research is focused on enhancing robustness at the expense of accuracy, much like optimizing for accuracy led to a lack of explainability. Similarly, scholars have studied the interplay with fairness as well as the possible issues stemming from it. These dimensions are not exclusive and need to be addressed holistically and considered on equal terms when aiming to build trustworthy and fair AI systems.

In this sense, sole data-driven approaches have shown their limitations. Discussions around these topics have pointed toward the need for integrating symbolic knowledge. However, few of them touched upon which kind of knowledge is needed and how to collect it. In subsection 7.2 and section 8, we provide a commentary on human-centered approaches and how these approaches can provide a path toward tackling the aforementioned challenges for robust AI.

## 7.2   Deepening the Research on Human Involvement for Existing Robustness Methods

A number of papers we surveyed implicitly involve humans to instantiate the methods they propose, either to assess or enhance a model's robustness. Yet, they do not delve deeper into the challenges for a human agent to perform their task, which constitutes an obstacle to the development of methods and frameworks for overcoming these challenges. This merits further investigation as such human involvement is essential to the success of the methods. Especially, we identify two main areas where human involvement is necessary but lacks research.

*7.2.1   Evaluating Robustness.* To design appropriate perturbations or attacks on which a model should be robust, one often needs human knowledge. For instance, Jin et al. [101] and La Malfa and Kwiatkowska [112] generate adversarial attacks on text samples, that have to verify a number of human-defined constraints for them to be deemed realistic by humans. Yet, designing such constraints and empirically evaluating (through user studies) to what extent the samples transformed by the corresponding constrained attack align with the human idea of "realistic" sample, has not been investigated extensively, despite how crucial that is for engineering "good" attacks.

In a similar fashion, works on robustness to natural perturbations should ideally define a comprehensive set of domain-specific perturbations relevant to the problem at hand and its context. However, to the best of our knowledge, existing works that develop benchmarks or robustness-enhancing methods [88, 108] with regard to such perturbations have not investigated ways to be more comprehensive. While we believe in the impossibility to reach comprehensiveness (previously unheard-of perturbations can always arise), one could develop tools to support the definition of relevant perturbations. For instance, we envision the usefulness of fine-grained, actionable taxonomies of perturbations (e.g., Koh et al. [108] talk about subpopulation shifts and domain generalization, but this might vary in different domains and types of tasks); collaborative documentation of domain-specific perturbations; libraries to generate such perturbations semi-automatically; and frameworks and metrics to uncover new types of perturbations in the wild, potentially involving humans in the runtime.

*7.2.2 Increasing Robustness.* Various methods that aim at increasing robustness implicitly employ humans, without extensive focus. Jin et al. [101], for instance, collect potential adversarial examples by executing a sequence of engineered steps, that could be refined by the practitioner who would leverage existing tools for, e.g., identifying synonyms and antonyms, ranking word importance, etc. Peterson et al. [165], Chang et al. [41], Nanda et al. [150], and Ning et al. [154] respectively show that one can train more robust models by leveraging human uncertainty on sample labels instead of using reconciled binary labels, by integrating human rationales for the labeling process into the training process, or by actively querying the most relevant levels of perturbations from an expert during training. While these are promising research directions, these works could further be improved by exploiting existing works on human computation assessing the quality of crowdsourced outputs [97], or designing crowdsourcing tasks that remove task ambiguity and lead to higher quality outputs [69], especially in the context of subjective tasks. This could serve to understand the nature of uncertainties and define rationales that are relevant to robustness.

# 8 A CONSPICUOUS ABSENT FROM THE LITERATURE: THE ML PRACTITIONER

Last but not least, our rigorous survey also reveals another prominent research gap: the absence of human-centered work in proposed approaches, and the lack of technologies and workflows to support ML practitioners in handling robustness. In this section, we discuss relevant research literature, and future research directions regarding this topic.

## 8.1 Robustness By Human-Knowledge Diagnosis

One of the most notable absentee from the retrieved papers is robustness by human-based diagnosis. Existing works focus on generating out-of-distribution data, in order to make a model fail, and later expose this model to this data during training to make it more robust. Especially for robustness to natural perturbations, this means that one should always characterize the type of data the model might encounter before being able to generate such data. This is not always possible in practice, e.g., due to contractual and privacy reasons, cost, temporal variability of contextual application of the model, etc. To circumvent this issue, a major, promising research direction surfaces from comparing the surveyed robustness methods to existing works in other computer science fields. This direction revolves around developing complementary, hybrid human-machine approaches, that would leverage research progress in human-centered fields, essentially explainability, crowdsourcing and human-in-the-loop machine learning (ML), as well as knowledge-based systems, to estimate model performance on more realistic data distributions without requiring such distributions.

*8.1.1 Existing Approaches.* Only few related works leverage human capabilities to identify and mitigate potential failures of a model. In particular, explanations for datasets [185] have been proposed, that could be leveraged by a practitioner to identify data skews that might impact the model performance. In this vein, Liu et al. [128] introduce a hybrid approach to identify unknown unknowns, where humans first identify and describe patterns in a small set of unknown unknowns, and then classifiers are learned to recognize these patterns automatically in new samples. Departing from datasets, Stacey et al. [209], and Arous et al. [11] have trained models whose features are better aligned with human reasoning (with the assumption that alignment leads to stronger robustness), by leveraging human explanations of the right answer to the inference task and controlling the features learned by the model during training to align with these human explanations.

*8.1.2   Envisioned Research Opportunity.* The above approaches reveal that instead of looking solely at the outputs of a model and its confidence in its predictions, one can leverage additional information such as the model features or training dataset, to estimate the model's robustness. Especially, even when a model prediction is correct, the model features might not be meaningful. Hence, assessing model features and their human-alignment can allow to shift from solely evaluating the correctness of the predictions on the available test, to indirectly assessing the robustness of the model to OOD data points. Moreover, understanding characteristics of the datasets that led to such learned features could later on serve to mitigate unaligned features.

*Surfacing Model Features using Research on Explainability and Human Computation.* To surface a model's features, one can rely on a plethora of explainability methods [185]. Certain models are built with the idea of being explainable by design [216, 274], while others are applied post-hoc interpretability methods [18, 176, 211], with different properties (e.g., different nature of explanations being correlation or causation -based, different scopes be it local or global, different mediums be it visual or textual, etc.) [126, 206]. It is now important to adapt such feature explanations to allow for checking their alignment with human expected features.

In that regard, the push towards human-centered explanations for ML practitioners is highly relevant. Existing explanations often leave space for many different human interpretations, for which the practitioners do not always have domain expertise to disambiguate the highest-fidelity features. For instance methods that output saliency maps [198] or image patches [77, 106] do not pinpoint to the actual human-interpretable features the model has learned. Yet, one might need clear human concepts to reason over the alignment of the features [17]. Hence, further research on *semantic, concept-based explanations* acquired via human computation is needed [18, 91].

*Leveraging Literature on Knowledge Acquisition for Identifying Expected Features.* To reason over feature alignment, one also needs to develop an understanding of the model expected features. While very few works have looked into this problem [193], existing works on commonsense-knowledge acquisition [266] could be leveraged to that end. These works propose to harvest knowledge automatically from existing resources such as text libraries, or through the involvement of human agents, e.g., through efficient and low-cost interactions within Games with a Purpose [16, 179, 234], or other types of carefully designed crowdsourcing tasks [96, 188]. One would need to investigate how to adapt such approaches to collect relevant knowledge, and how to represent this knowledge into relevant feature-based information.

*Comparing Features via Reasoning Frameworks and Interactive Tools.* Finally, practitioners need tools to check the alignment between the model and expected features. Interactive frameworks and user interfaces [17], e.g., *Shared Interest* [27], take a step in that direction as they enable manual exploration of model features, with various degrees of automation for comparing to expected features. Inspired by the literature on AI diagnosis, such as abductive reasoning [54, 178], automated feature-reasoning methods could also fasten the process while making it more reliable.

## 8.2   Involving Humans in Other Phases of the ML Lifecycle

Broader ML literature has also proposed other approaches to involve humans and make "better" models. Yet, none of these approaches has considered making the models more robust. Instead, they focus on increasing the performance of the model on the test set. Hence, we suggest to investigate how to adapt such approaches to increase model robustness.

*8.2.1    ML with a Reject Option.* While ML models typically make predictions for all input samples, this might not be reasonable and turn dangerous in high-stake domains, when the predictions are likely to be incorrect. Accordingly, a number of research works have developed methods to learn when to appropriately reject a prediction, and defer the decision about the sample to a human agent [87]. Proposed rejectors can either be *separate rejectors* placed before the predictor, that select the input samples to input to this predictor; *dependent rejectors* placed after the predictor and re-using its information (e.g., confidence metrics) to decide which predictions not to account for; and *integrated rejectors* that are combined to the predictor, by treating the rejection option as an additional label to the ones to predict. Each type of rejector bears advantages and disadvantages based on the context of the decision, and would merit being adapted to robustness, as we only found few works towards that direction [105, 160, 212].

*8.2.2    Human-in-the-Loop ML Pipelines.* Human-in-the-Loop (HIL) ML [230] is traditionally concerned with developing learning frameworks that account for the noisy crowd labels [174], or "learning from crowds", through models of the annotation process (e.g., task difficulty, task subjectivity, annotator expertise, etc.). Such frameworks often rely on active learning to reduce annotation cost [254, 256]. More recent works around HIL ML also devise new approaches to build better model pipelines by involving the crowd, such as to identify weak components of a system [158], to identify noise and biases in the training data [95, 257], or to propose potential data-based explanations to wrong predictions [34]. While we could find a few works that investigate the intersection between active learning and adversarial training [139, 141, 197, 203], we could not find any work that looks more broadly at the different types of robustness, and the different ways of bringing humans in the ML pipeline. These intersections are yet promising as they constitute more realistic scenarios of the development of ML systems and they succeeded in making models more accurate in the past.

## 8.3    Supporting ML Practitioners in Handling Robustness

Looking beyond the research world towards the practice, it is always an ML practitioner who builds the ML system. Hence, it is not sufficient to develop methods that can work in theory, but it is also important to understand the obstacles practitioners actually encounter in making their systems robust. While studying the gap between research and practice has revealed highly insightful in the past for various ML contexts [92, 93, 110, 127, 168], to the best of our knowledge, it has not been studied in the context of ML robustness. Possibly the closest work is the interview study of Shankar et al. [190] that investigated MLOps practices beyond the development of a model towards production and monitoring of data shifts or attacks.

*8.3.1    Understanding Practices Around Robustness.* The human-computer interaction community (HCI) has performed qualitative, empirical, studies, typically based on semi-structured interviews with ML practitioners, to understand how these practitioners build ML models with certain considerations in mind. These considerations revolve around the different steps practitioners take, e.g., challenges of collaboration for each step [110, 168], and the use in certain of these steps of tools such as explainability methods [92, 93, 127] or fairness toolkits [117, 177]. These studies have resulted in frameworks modeling the practitioner's process, lists of challenges, and discussions around the fit of existing methods and tools to answer these challenges. We argue that adopting similar research questions and methodologies (e.g., semi-structured interviews with hypothetical scenarios or practitioner's own tasks, ethnographies, etc.) would also reveal useful to better direct robustness research in the future. For instance, Liao et al. [127] have constituted an explainability question bank that highlights the questions practitioners ask when building a model by exploiting explainability, and that can serve to identify research opportunities through questions still difficult to answer. A robustness question bank would similarly provide a structured understanding of what is still lacking. Moreover, HCI research investigating practices around ML fairness [56] has shown a major gap in terms of guidance for practitioners to choose appropriate fairness metrics and mitigation methods. Acknowledging the plethora of robustness metrics and methods, we envision that user-studies around robustness would reveal a similar gap, that could be filled by taking inspiration from the fairness literature.

*8.3.2   Integrating Robustness into Existing Workflows.* Some works have also focused on developing workflows and tools to support practitioners in model building. These works often revolve around user interfaces to more easily investigate a model and its training dataset, and identify failures or bugs [17, 151]. Other works build tools, e.g., documentation or checklists, [7, 29, 74, 144] and workflows [205] to support making and documenting relevant choices when building or evaluating a model. We argue that robustness research should not only focus on algorithmic evaluation and improvement, but also aim at developing new supportive tools and integrating them into existing solutions. In relation to that, and possibly closest to supporting practitioners in handling robustness, Shen [195] propose the idea of establishing trust contracts, i.e., contract data distributions and tasks that define the type of task and data that is in- and out-of-distribution. Yet, this remains challenging as there is no appropriate way to formalize such contracts.

## 9   CONCLUSION

In this survey, we collected, structured, and discussed literature related to robustness in AI systems. To this end, we performed a rigorous data collection process where we collected, filtered, summarized and organized literature related to AI robustness generated in the last 10 years. Based on this literature, we first discussed the main concepts, definitions, and domains associated with robustness, disambiguating the terminology used in this field. We then generated a taxonomy to structure the reviewed papers and to spot recurring themes. We identified three main themes and thoroughly discussed them. In particular, we focused on (1) fundamental approaches to improve model robustness against adversarial and non-adversarial perturbations, (2) applied approaches to enhance robustness in different application areas, and (3) evaluation approaches and insights. We finalized our paper by describing the research gaps identified in the literature and by detailing future lines of work that involve including humans as central actors. We argued that humans could play a fundamental role in improving, evaluating, and validating AI robustness. In conclusion, we contributed to the exiting literature with an informative review that summarizes and organizes recent work in the field of AI robustness, while it also suggests novel human-centered approaches for the research community to explore, discuss, and further develop.

## 10   ACKNOWLEDGEMENTS

This research has been partially supported by the TU Delft *Design@Scale* AI Lab, by the *HyperEdge Sensing* project funded by Cognizant, by the European Commission under the H2020 framework, within project 101016233 PERISCOPE (Pan-European Response to the ImpactS of COVID-19 and future Pandemics and Epidemics), by the European Union's Horizon 2020 research and innovation programme under the Marie Skłodowska-Curie grant agreement No 955990, and by the Ph.D. Scholarship on Explainable AI funded by Cefriel.

# Supplementary Material for *A.I. Robustness: a Human-Centered Perspective on Technological Challenges and Opportunities*

ANDREA TOCCHETTI*, LORENZO CORTI*, AGATHE BALAYN*, MIREIA YURRITA, PHILIPP LIPPMANN, MARCO BRAMBILLA, and JIE YANG†

This supplementary material consists of additional commentary on the literature we surveyed in the main manuscript *A.I. Robustness: a Human-Centered Perspective on Technological Challenges and Opportunities*, with which it shares the same structure. In particular, additional content for Sections 4, 5, and 6 is reported here.

## 1 SUPPLEMENT FOR SECTION 4: ROBUSTNESS BY

### 1.1 Processing the Training Data

*1.1.1 Augmenting Data for Adversarial Robustness.* Most of the identified literature focuses on transforming [38, 66, 95], generating [7, 31, 32, 38, 49, 73, 74, 99] or employing ready-to-use [21] data and/or adversarial samples to extend or create datasets to train more robust models. Such a data augmentation process can successfully improve adversarial robustness [38, 66, 73, 95, 99, 113], adversarial accuracy [1], and fairness [88] while sometimes reducing time costs [14], and adversarial attack success rate [10].

*1.1.2 Augmenting Data for Non-Adversarial Robustness.* While addressing model robustness, researchers also focus on improving models' performance [40, 54, 69, 75] and noise robustness [40, 54, 75] through data augmentation techniques. In particular, common perturbations [40], GANs [75], and data-driven approaches [54, 69] are employed to generate data to expand the training data, consequently achieving improved robustness [40, 54, 75] and accuracy [40, 54, 69, 75]. Similarly, researchers demonstrate that data augmentation can also benefit model generalization [47, 73, 99, 112]. However, augmenting datasets by generating new samples is not the only viable approach. Sometimes, it is enough to use similar data, e.g., out-of-distribution data from other datasets [29] or unlabeled data [107]. Furthermore, it has been proven that noise can also be applied to generate data that cannot be learned through adversarial training, preventing attackers from exploiting the targeted model [26].

---

*The authors contributed equally to this research.

†Andrea Tocchetti and Marco Brambilla are with Politecnico di Milano, Email: {andrea.tocchetti, marco.brambilla}@polimi.it; Lorenzo Corti, Agathe Balayn, Mireia Yurrita, Philip Lippmann, and Jie Yang (corresponding author) are with Delft University of Technology, Email: {l.corti, a.m.a.balayn, m.yurritasemperena, p.lippmann, j.yang-3}@tudelft.nl.

---

*1.1.3   Frequently Used Datasets.* When it comes to the datasets researchers rely on when studying robustness, we identify: MNIST [1, 7, 10, 14, 29, 38, 47, 49, 57, 68, 89, 92, 98, 99, 107, 108, 112] and Fashion-MNIST [14, 89], CIFAR-10 [7, 14, 21, 26, 27, 32, 38, 47, 49, 68, 73, 89, 98, 99, 107, 108, 113] and its variants [21, 26, 47, 112], SVHN [21, 27, 29, 38, 57, 98], and ImageNet [26, 44, 49, 66] and its variants [14, 21, 44, 47, 68, 110]. Such findings highlight the broad interest of the research community in the robustness of computer vision models.

## 1.2   Designing In-Model Robustness Strategies

### 1.2.1   Training for Robustness.

*Training with Dynamic Perturbations.* Differently from the majority of existing approaches, instead of perturbing data instances, Hosseini et al. [35] work by applying random subsampling and training neural networks on different subsets of pixels.

*Training Robust Feature Representations.* Scholars have also designed novel methods to learn more robust feature representations [16, 50, 61, 76, 111]. Connected to this idea, Shu et al. [85] explore perturbations of feature statistics based on the magnitude of their effect, while Eigen and Sadovnik [23] propose to nudge the output to be dependent on the $k$-largest network activations. Finally, Chen and Lee [17] obtain robust features through the addition of an auxiliary model. The objective is to help the original model to learn features even when subject to perturbations.

*Adversarial Training Algorithms.* In this context, researchers have proposed a plethora of algorithms [3, 44, 67, 86, 93], also borrowing from different Machine Learning paradigms like Self-Supervised [67] and Unsupervised learning [93], that are applicable to a variety of tasks. Connected to leveraging input spaces, Liu and Lomuscio [59] propose a black-box training method that explores small regions around input instances that are more likely to lead to stronger adversaries. Instead, when it comes to a model's internals, Rozsa and Boult [79] take a more fine-grained approach and, motivated by the open space problem of activation functions such as ReLU, use Tent activation functions to reduce the neurons' output surface exploitable by an attacker. Within the same context, Pereira et al. [72] identify the best layers in Transformer architecture to perturb during fine-tuning. Targeting the late layers of neural networks, Antonello and Garner [4] use *t-softmax*, a novel operator based on t-distribution, which can better describe uncertainty inherent to out-of-distribution data and attribute them low confidence values. On the same note, Kwon and Lee [42] also deal with manipulating confidence values and suggest a methodology for resisting adversarial attacks by providing random confidence values of the output. Finally, Sengupta et al. [82], argue that non-robust models tend to rely on features that humans would not consider, and propose the use of ground truth labels to degrade performance with respect to a human observer.

*Training with Adapted Regularizers.* Among the regularizers discussed in the main manuscript, Xu et al. [103] suggest a consistency-based regulariser that keeps model predictions stable in the neighborhood of misclassified adversarial examples. On the other hand, Xu et al. [102] design a regularisation term based on increasing the angular margin of weight vectors of a classifier. Finally, Vinh et al. [96] tackle regularization from a different angle. They demonstrate that perturbed mini-batches obtained through Random Projection can produce robust and regularised models.

*Beyond Adversarial Training.* Amid alternative approaches to adversarial training, Serban et al. [83] argue for training models through prototypes which lead to *inter-class separability* and *intra-class compactness*.

### 1.2.2   Designing Robust Architectures.

*Tweaking Neural Network Layers.* In this context, Barros and Barreto [9] introduce M-estimators, a widely used parameter estimation framework in robust regression, to compute the output weight matrix and deal with label noise. Furthermore, Lecuyer et al. [45] add a (zero-mean) noise layer between every other layer in the network. On a slightly different note, Baidya et al. [8] seek to bridge models and the biological concept of vision by applying VOneNet [20], a Convolution-based network simulating humans' primary visual cortex, as a robust feature extractor for existing Computer Vision models for image classification. While all the aforementioned methods result in overall better-performing models, they are very specific and selected implementations. Finally, related to differential equations-inspired neural networks, Liu et al. [56] have incorporated stochastic regularizations into Neural Ordinary Differential Equations to improve their robustness on image classification tasks while keeping stable both adversarial and non-adversarial performance. Li et al. [51] leverage the numerical stability of implicit Ordinary Differential Equations and propose Implicit Euler skip connections (IE-Skips) by modifying the original skip connection in ResNet.

*Searching Neural Architectures.* As an extension to Neural Architecture Search, Kotyan and Vargas [41] propose Robust Architecture Search (RAS): an evolution of NAS which uncovers inherently robust networks by evaluating layers and blocks in terms of the number of parameters and models in terms of adversarial robustness.

### 1.3 Leveraging Model Post-Processing Opportunities

*1.3.1 Identifying Unnecessary or Unstable Model Attributes (neurons, features).* Raviv et al. [77] deal with unstable neurons and suggest a novel Fourier stabilization approach to replace the weights of individual neurons with robust analogs derived using Fourier analytic tools. Furthermore, pruning has also been tested and shown to be effective in improving certified robustness [53]. On the premise of inspecting the effectiveness of different pruning methods, Liu et al. [60] propose a visual technique for such a task.

*1.3.2 Fusing Models.*

*Against Input issues.* Additional methods for achieving post-model-training robustness through auxiliary models identifying adversarial data instances are the ones by Metzen et al. [65] and Akumalla et al. [2].

*Against Infected Models.* Wu et al. [101], for example, combine the usage of a classifier built on top of a neural network that aimed at identifying patterns in hidden unit activations with a strategy that uses the lack of stability to weight changes of wrong predictions to differentiate them from right predictions.

*Improving Fusion Models' own Robustness.* In addition to fusing models to enhance robustness, the robustness of fusion models themselves has also been studied. For example, Khalid and Arshad [39] study label noise in ensemble classifiers. Another work focusing on ensemble models is that of Pang et al. [70]. They suggest a new definition of ensemble diversity as the diversity among non-maximal predictions of individual members and used this definition to present an Adaptive Diversity Promoting (ADP) regularizer. This regularizer improves the robustness of the ensemble by making adversarial examples difficult to transfer among individual members. In a similar fashion, Goldblum et al. [30] study the transferability of adversarial robustness from teacher to student during knowledge distillation and find that knowledge distillation is able to preserve much of the teacher's robustness to adversarial attacks even without adversarial training for most datasets.

## 2 SUPPLEMENT FOR SECTION 5: ROBUSTNESS FOR

### 2.1 Robustness for Specific Application Areas

*2.1.1 Robustness for Natural Language Processing (NLP) Tasks.* Nowadays, fine-tuning pre-trained models is a widespread practice. To that aim, Pereira et al. [72] propose an enhanced adversarial training algorithm for fine-tuning transformer-based language models by identifying the best combination of layers to add the adversarial perturbation. On the other hand, Du et al. [22] take advantage of the adversarial examples generated through Probabilistic Weighted Word Saliency [78] for training purposes.

*Robustness for Machine Translation.* A particular focus for NLP robustness research has been the domain of machine translation, where small perturbations in the input can severely distort intermediate representations and thus impact the final translation output. Such perturbations, or noise, can either be naturally occurring or synthetic. Synthetic noise, which is easier to control and obtain is used by Vaibhav et al. [94] to enhance the robustness of MT systems by emulating naturally occurring noise in otherwise clean data. They are thus able to make a translation system more robust to naturally occurring noise in the test set by including synthetic noise in the training data. Another data augmentation approach is proposed by Li and Specia [54] who find that the use of noisy parallel data can improve model robustness on noisy and clean datasets alike. Further, they observe that the introduction of external data with different types of noise may improve the model's robustness more generally, even without the usage of in-domain data.

*2.1.2 Robustness for Cybersecurity.* As Machine Learning techniques are being applied in Cybersecurity, the robustness of such systems to malicious actors becomes more and more of a concern. Patil et al. [71] use adversarial training on existing malware detection deep learning techniques. Here, they report improvements in robustness across all considered architectures following the retraining using adversarial examples. Finally, Melis et al. [64] study the usability of gradient-based attribution methods to identify more robust algorithms. They find a connection between the evenness of explanations and adversarial robustness.

## 3 SUPPLEMENT FOR SECTION 6: ROBUSTNESS ASSESSMENT AND INSIGHTS

### 3.1 Evaluation procedures

*3.1.1 Evaluation Strategies.*

*Evaluation of Robustness.* Besides studying how to devise defense mechanisms, it is fundamental to define evaluation criteria for robustness. Arcaini et al. [6] frame robustness as the model's capability to correctly classify perturbed data from multiple perturbed datasets. Another approach proposed by Lim et al. [55] is to evaluate pointwise $L_p$-Robustness by checking all the activation regions around a particular data point.

*Evaluation of Certified Robustness.* When it comes to the assessment of certified robustness, most of the literature focuses on model robustness against adversarial attacks and samples [11] such that end-users can act exclusively on robust model predictions.

*3.1.2 Benchmarks.* In the context of benchmarking datasets, Lee et al. [46] extend the Schema-Guided Dialogue dataset to measure the robustness of dialogue systems to linguistic variations.

*3.1.3 Metrics.*

*Metrics for Adversarial Robustness.* Metrics for quantifying adversarial robustness have been introduced. Another example of these is the one by Buzhinsky et al. [12], tailored to robustness against adversarial attacks, under the lenses of different interpretations.

*Metrics for Adversarial Attacks.* Despite the necessity for metrics to assess model robustness, other metrics prove useful in evaluating different aspects concerning robustness [43], like its relationship with adversarial examples and accuracy, or whether a generated architecture is statistically accurate [43]. More specifically, Arcaini et al. [5] propose a metric to evaluate the robustness of models when images are subject to acquisition alterations.

In general, metrics are generated by capturing the decision boundary of deep neural networks by the distribution of the data concerning such boundaries [18].

*Metrics for the Complexity of Robustness Methods.* While most researchers focus on implementing evaluation approaches, a small research branch focuses on improving their efficiency, mainly enhancing their precision in computing the robustness bounds and reducing their computational complexity and execution time [34, 48, 97].

## 3.2 Studies around Proposed Robustness Methods & Insights

### 3.2.1 Insights on Adversarial Robustness.

*Assessment Methods and Frameworks.* Recent studies have suggested methodologies and frameworks for assessing robustness of AI models against adversarial attacks [15, 24, 37, 80, 115]. For example, Chang et al. [15] describe an attack-agnostic method to assess the robustness of AI models. Instead, Zimmermann et al. [115] propose an active robustness test to identify weak attacks leading to weak adversarial defense evaluations. Finally, Sehwag et al. [80] analyze the robustness of open-world learning frameworks in the presence of adversaries by means of out-of-distribution adversarial samples. They also outline a preliminary solution for such a problem which, besides better robustness to threats, enables trustworthy detection of out-of-distribution inputs. Fawzi et al. [24] describe a theoretical framework to analyze the robustness of binary classifiers to adversarial perturbations and show fundamental upper bounds on the robustness of classifiers. Mahima et al. [62] propose an approach where a model is subject to adversarial perturbations and physical image corruptions. Here, robustness is measured as the ratio between the accuracies of corrupted input to clean input. They find that adversarial examples and physical distortions seem to cause the networks to attend more sparsely to different parts of an image.

*Based on Comparisons.* Beyond formal methods and frameworks, there are several examples of papers empirically evaluating robustness through comparison [13, 19, 52, 58, 84, 91, 100, 109]. Zhang et al. [109] test whether ensemble classifiers are more robust than single classifiers in cases where attackers can only get a portion of the labeled data. They find that ensemble classifiers are not necessarily more robust, as they are more susceptible to evasion attacks. Carrara et al. [13] analyze the robustness of image classifiers implemented with Ordinary Differential Equation networks against adversarial attacks and compare them with standard deep models. Chun et al. [19] empirically evaluate the uncertainty and robustness of different image classifiers that have been trained using regularization methods. Sharmin et al. [84] analyze the robustness of bio-plausible networks, i.e., spiking neural networks, under adversarial tests compared to VGG-9 artificial neural networks, and conclude that spike neural networks are more robust than artificial neural networks. Li et al. [52] compare the robustness of discriminative and generative classifiers, i.e., deep Bayes classifiers, against adversarial attacks. Liu et al. [58] study the impact of adversarial attacks against a convolutional LeNet-5 to observe the changing law of the adversarial robustness of the deep learning model. Tarchoun et al. [91] investigate the effect on robustness that view angle has in multi-view datasets when employing adversarial patches. Wu et al. [100] examine the relationship between network width and model robustness for adversarially trained neural networks and find that robustness is closely related to the trade-off between accuracy and perturbation stability.

*Based on the Investigation of Activation Function and Weights Perturbations.* Su et al. [87] evaluate the robustness of activation functions in convolutional neural networks through the possible differences in attack confidence when deep convolutional networks use sigmoid compared to ReLU activation functions. They find that sigmoid functions cause the attacker's confidence to be smaller.

### 3.2.2 Insights on Robustness to Natural Perturbations.

*About robustness to Noise.* A prominent line of work is constituted by studies evaluating the robustness of AI systems against noise [25, 28, 104]. Ghosh et al. [28] study the performances of commonly used convolutional neural networks against image degradations, such as Gaussian noise and blur. They propose a novel, degradation adaptive method to improve the performances of such networks when degradation is present in the input data. Franceschi et al. [25] also study robustness to Gaussian and uniform noise and characterize this robustness to noise in terms of the distance to the decision boundary of the classifier. Another study on label noise, conducted by Xue et al. [104], investigate why contrastive learning leads to improved robustness against label noise.

*About Robustness to Differences in Distributions.* Another area of interest within the evaluation of robustness to natural perturbation is that of differences in distribution [81, 90, 114]. Sengupta and Friston [81] study the robustness of deep neural networks by evaluating the robustness of three recurrent neural networks to tiny perturbations, on three widely used datasets, to argue that high accuracy does not always mean a stable and robust system. There have also been multiple works on evaluating neural networks' robustness to data shifts and out-of-distribution data [90, 114].

### 3.2.3 Robustness Evaluation in Computer Vision. 
Robustness evaluation efforts have also focused on specific architectures within Computer Vision. Mathis et al. [63] evaluate the robustness of ImageNet-performing architectures on out-of-domain training data and find that performance on ImageNet predicts generalization for within- and out-of-domain data on the task of interest, hence demonstrating robustness. Zaidi et al. [106] analyze the conditions in which invariances to picture transformations emerge in deep convolutional neural networks, and subsequently demonstrate that increasing the amount of seen-transformed examples increases both the invariance to transformations and the robustness to transformations of unseen-transformed categories. Yin et al. [105] study how robustness is influenced by different types of perturbations, and whether there are trade-offs with respect to other observable variables. When it comes to specific computer vision tasks, the evaluation of the robustness of face-recognition approaches has proliferated [33]. Besides, Huang et al. [36] present a framework for the analysis of the robustness of visual question-answering models.